\documentclass[10pt,twocolumn,letterpaper]{article}

\usepackage{iccv}
\usepackage{times}
\usepackage{epsfig}
\usepackage{graphicx}
\usepackage{amsmath}
\usepackage{amssymb}

\usepackage{color}
\usepackage{setspace}

\usepackage{multirow}
\usepackage{booktabs}

\usepackage{diagbox}
\usepackage{rotating}
\usepackage{overpic}
\usepackage{booktabs}

\usepackage{graphicx}
\usepackage{caption}
\usepackage{subcaption}

\usepackage{array}
\newcolumntype{x}[1]{>{\centering\arraybackslash\hspace{0pt}}p{#1}}
\newcolumntype{t}[1]{>{\centering\arraybackslash\hspace{0pt}}p{#1}}

\usepackage{sty/algorithm}
\usepackage{sty/algorithmic}

\usepackage{pifont}
\newcommand{\cmark}{\ding{51}}%
\newcommand{\xmark}{\ding{55}}%

\graphicspath{{./figures/exampleSegmentations/}}

\newcommand{\argmax}{\operatornamewithlimits{argmax}}

\newlength{\Oldarrayrulewidth}

\newcommand{\figref}[1]{Fig.~\ref{#1}}

\newcommand{\equref}[1]{Eqn.~(\ref{#1})}
\newcommand{\secref}[1]{Sec.~\ref{#1}}

\usepackage[pagebackref=true,breaklinks=true,letterpaper=true,colorlinks,bookmarks=false]{hyperref}

\iccvfinalcopy 

\def\iccvPaperID{****} 

\ificcvfinal\fi

\begin{document}

\title{Bottom-Up Top-Down Cues for Weakly-Supervised Semantic Segmentation}

\author{Qibin Hou$^{1}$\thanks{Joint first author}\\
\and
Puneet Kumar Dokania$^{2}$\footnotemark[1]\\
\and
Daniela Massiceti$^{2}$\\
\and
Yunchao Wei$^{3}$\\
\and
Ming-Ming Cheng$^{1}$\\
\and
Philip H. S. Torr$^{2}$\\
\and
\\$^1$CCCE, Nankai University \; \; $^2$ University of Oxford \; \; $^3$NUS\\
}

\maketitle
\pagestyle{plain}
\pagenumbering{arabic}

\begin{abstract}
We consider the task of learning a classifier for
semantic segmentation using weak supervision in the form of
image labels which specify the object classes present in the image.
Our method uses deep convolutional neural networks ({\sc cnn}s)
and adopts an Expectation-Maximization (EM) based approach.
We focus on the following three aspects of
EM: (i) initialization;
(ii) latent posterior estimation (E-step) and
(iii) the parameter update (M-step).
We show that saliency and attention maps, our bottom-up and top-down cues respectively, of simple images provide very good cues to learn an initialization for the EM-based algorithm. Intuitively, we show that before trying to learn to segment complex images, it is much easier and highly effective to first learn to segment a set of simple images and then move towards the complex ones.
Next, in order to update the parameters,
we propose minimizing the combination of the standard
\textit{softmax} loss and the KL divergence between the
true latent posterior and the likelihood given by the {\sc cnn}.
We argue that this combination is more robust to wrong
predictions made by the expectation step of the EM method.
We support this argument with empirical and visual results.
Extensive experiments and discussions show that:
(i) our method is very simple and intuitive;
(ii) requires only image-level labels; and
(iii) consistently outperforms other weakly-supervised
state-of-the-art methods with a very high margin
on the PASCAL VOC 2012 dataset.
\end{abstract}
\newcommand{\addSubFig}[3]{\begin{subfigure}[t]{.15\linewidth}
   \includegraphics[width=\linewidth]{./figures/mining/#1}
   \caption{#2}\label{#3}\end{subfigure}
}
\begin{figure*}[ht!]
    \centering
    \addSubFig{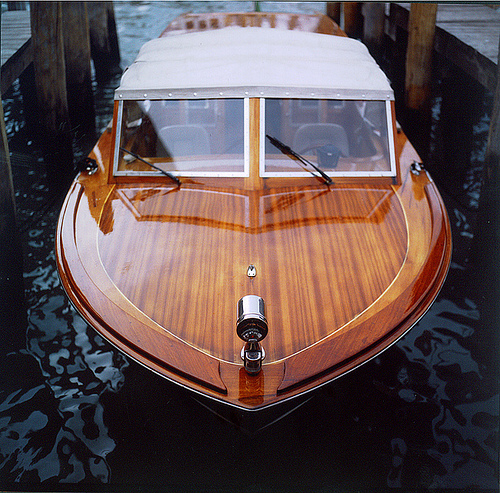}{Input Image}{fig:origImage} ~
    \addSubFig{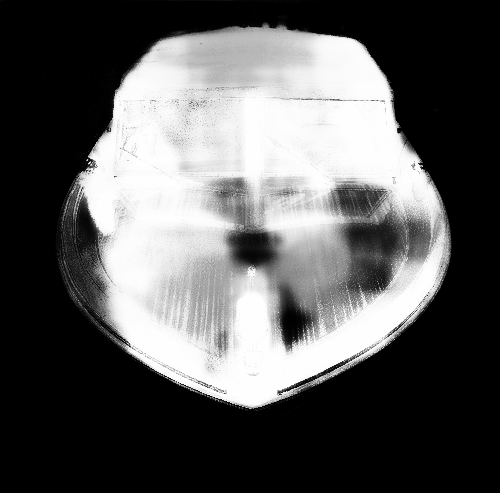}{Saliency Map}{fig:imgSal} ~
    \addSubFig{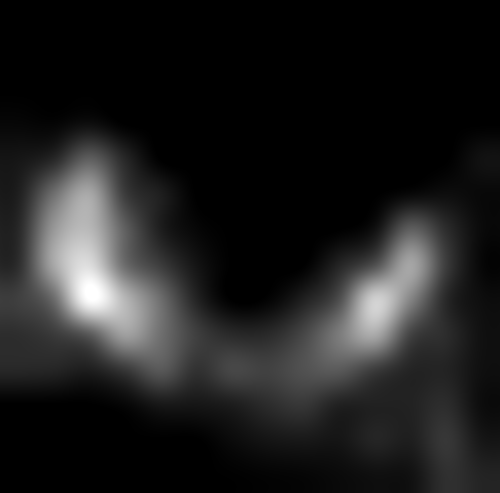}{Attention Map}{fig:imgAtt} ~
    \addSubFig{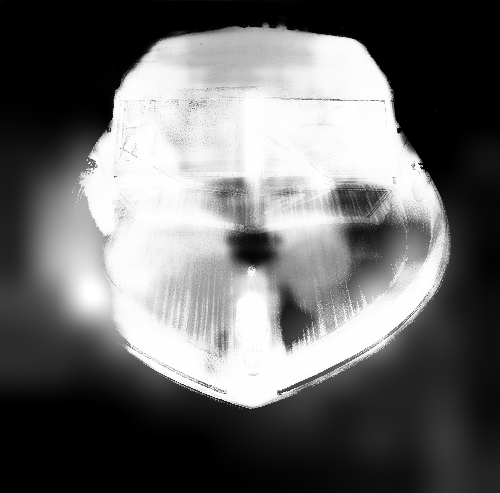}{Combined Map}{fig:minedPixels}
    \caption{An example of combining bottom-up and top-down cues for simple images. (a) Input image with `boat'. (b) Saliency map~\cite{Hou2016DeepSaliency} (`bottom-up' cue). (c) Attention map~\cite{Zhang16Attention} (`top-down' cue). (d) Combined map - notice that the saliency map puts high probability mass on regions missed by the attention map, and vice versa.
    }\label{fig:pixelMining}
\end{figure*}
\vspace{-6mm}
\section{Introduction}
The semantic segmentation task has rapidly advanced with the use of Convolutional Neural Networks
(\textsc{cnn}s)
\cite{Chandra16FastExactSemantic,Chen15DeepLabV1,Long15FCN,Zheng15CRFRNN}.
The performance of \textsc{cnn}s, however,
is largely dependent on the availability of a large corpus
of annotated training data,
which is both cost- and time-intensive to acquire.
The pixel-level annotation of an image in PASCAL VOC takes
on average $4$ minutes~\cite{Bearman16Point},
which is likely a conservative estimate given that it is
based on the COCO dataset \cite{Lin14COCO}
in which ground-truths are obtained by annotating
polygon corners rather than pixels directly.
In response, recent works focus on weakly-supervised
semantic segmentation
\cite{Bearman16Point,Kolesnikov16SEC,Papandreou15EM,Pathak15CCNN,Pinheiro15Image2Pixel,Qi16AugFeedBack,Yunchao15STC}.
These differ from fully-supervised cases in that rather
than having pixel-level ground-truth segmentations,
the supervision available is to some lesser degree (image-level labels
\cite{Kolesnikov16SEC,Papandreou15EM,Pathak15CCNN,Pinheiro15Image2Pixel}, bounding boxes~\cite{Papandreou15EM},
or points and scribbles
\cite{Bearman16Point,Lin16ScribbleSup,Xu15VariousWeak}).

In this work, we address the semantic segmentation task using only image labels,
which specify the object categories present in the image.
Our motivation for this is two-fold:
(i) the annotation of an image with $20$ object classes
in PASCAL VOC is estimated to take $20$ seconds,
which is at least $12$ times faster than
a pixel-level annotation and is also scalable,
and (ii) images with their image labels or tags
can easily be downloaded from the Internet,
providing a rich and virtually infinite source
of training data.
The method we adopt, similarly to the weakly-supervised
semantic segmentation of \cite{Papandreou15EM},
takes the form of Expectation-Maximization
(EM)~\cite{Dempster77EM,McLachlan97EMBook}.
An EM-based approach has three key steps:
(i) initialization;
(ii) latent posterior estimation (E step);
and (ii) parameter update (M step).
We focus on all of these aspects. In what follows we briefly talk about each of them.

We provide an informed initialization to the EM algorithm by training an initial model for the semantic segmentation task using an approximate ground-truth obtained using the combination of class-agnostic
{\em saliency} map~\cite{Hou2016DeepSaliency} and class-specific
{\em attention} maps~\cite{Zhang16Attention} on a set of simple images with one object category (\textit{ImageNet}~\cite{Deng09ImageNet} classification dataset). This intuition comes from the obvious fact that it is easier to learn to segment simple images and then progress to more complex ones, similar to the work of \cite{Yunchao15STC}. Given an image, our saliency map {\em finds} the object (Figure~\ref{fig:imgSal}) - this is a class-agnotic `bottom-up' cue. Added to this, once provided with the class present in the image (from the image label - `boat' in this case), our attention map (Figure~\ref{fig:imgAtt}) gives the `top-down' class-specific regions in the image. Since both saliency and attention maps are tasked to find the same object, their combination is more powerful than if either one is used in isolation, as shown in Figure~\ref{fig:minedPixels}. The combined probability map is then used as the ground-truth for training an initial model for the semantic segmentation task. The trained initial model provides the initialization parameters for the follow-up EM algorithm. Notice that this initialization is in contrast to~\cite{Papandreou15EM} where the initial model is trained for the {\em image classification task} on the same \textit{ImageNet} dataset. Experimentally we have found that this simple way of combining bottom-up with top-down cues on the \textit{ImageNet} dataset (with \textit{no} images from PASCAL VOC 2012) allows us to obtain an initial model capable of outperforming all current state-of-the-art algorithms for the weakly-supervised semantic segmentation task on the PASCAL VOC 2012 dataset. {\em This is surprising since these algorithms are significantly more complex and they rely on higher degrees of supervision such as bounding boxes, points/squiggles and superpixels.} This clearly indicates the importance of learning from simple images before delving into more complex ones. With the trained initial model, we then incorporate PASCAL VOC images (with multiple objects) for the E and M Steps of our EM-based algorithm.

In the E-step, we obtain the latent posterior probability distribution by constraining (or regularizing) the {\sc cnn} likelihood using image labels based prior. This reduces many false positives by redistributing the probability masses (which were initially over the $20$ object categories) among only the labels present in the image and the backgroud. In the M-step, the parameter update step, we then minimize a combination of the standard \textit{softmax} loss (where the ground-truth is assumed to be a Dirac delta distribution) and the KL divergence~\cite{Kullback51KLdivergence} between the latent posterior distribution (obtained using the E-step) and the likelihood given by the {\sc cnn}. In the weakly-supervised setting this makes the approach more robust than using the \textit{softmax} loss alone since in the case of confusing classes,
the latent posterior (from the E-step) can sometimes
be completely wrong.
In addition to this, to obtain better {\sc cnn} parameters,
we add a probabilistic approximation of the
Intersection-over-Union (IoU)
\cite{Ahmed15ExpectedIoU,Cogswell14IoUCNN,Nowozin14IoU}
to the above loss function.

With this intuitive approach we obtain
state-of-the-art results in the weakly-supervised
semantic segmentation task on the PASCAL VOC 2012 benchmark
\cite{Everingham14VOC12}. We beat the best method which uses image label supervision by 10\%.
\section{Related Work}
Work in weakly-supervised semantic segmentation has explored varying levels of supervision including combinations of image labels
\cite{Kolesnikov16SEC,Papandreou15EM,Pathak15CCNN,Yunchao15STC},
annotated points~\cite{Bearman16Point}, squiggles~\cite{Lin16ScribbleSup,Xu15VariousWeak},
and bounding boxes \cite{Papandreou15EM}.
Papandreou \etal~\cite{Papandreou15EM} employ an EM-based approach with supervision from image labels and
bounding boxes. Their method iterates between inferring a latent
segmentation (E-step) and optimizing the parameters
of a segmentation network (M-step) by treating the
inferred latents as the ground-truth segmentation.
Similarly, \cite{Yunchao15STC} train an initial network using saliency maps,
following which a more powerful network is trained using
the output of the initial network.
The {\sc mil} frameworks of \cite{Pinheiro15Image2Pixel}
and \cite{Pathak14MIL} use fully convolutional networks
to learn pixel-level semantic segmentations from only
image labels.
The image labels, however, provide no
information about the position of the objects in an image.
To address this, localization cues can be used
\cite{Pinheiro15Image2Pixel,Qi16AugFeedBack},
obtained through indirect methods like bottom-up proposal
generation (for example, {\sc mcg}
\cite{Arbelaez14MCG}),
and saliency- \cite{Yunchao15STC} and attention-based
\cite{Zhang16Attention} mechanisms.
Localization cues can also be obtained directly through
point- and squiggle-level annotations
\cite{Bearman16Point,Lin16ScribbleSup,Xu15VariousWeak}.

Our method is most similar to the EM-based
approach of~\cite{Papandreou15EM}.
We use {\em saliency} and {\em attention} maps to learn
a network for a simplified semantic segmentation task which
provides better initialization of the EM algorithm.
This is in contrast to~\cite{Papandreou15EM}
where a network trained for a classification task is
used as initialization.
Also different from ~\cite{Papandreou15EM}
where the latent posterior is approximated by a
Dirac delta function (which we argue is too harsh
of a constraint in a weakly-supervised setting),
we instead propose to use the combination of the true
posterior distribution and the Dirac delta function
to learn the parameters.

\section{The Semantic Segmentation Task}
\label{sec:semSeg}
Consider an image $I$ consisting of a set of pixels
$\{ y_1, \cdots, y_n \}$.
Each pixel can be thought of as a random variable
taking on a value from a discrete semantic label set
$\mathcal{L} = \{l_0, l_1, \cdots, l_c\}$,
where $c$ is the number of classes ($l_0$ for
the background).
Under this setting, a semantic segmentation is defined
as the assignment of all pixels to their
corresponding semantic labels, denoted as ${\bf y}$.

{\sc cnn}s are extensively used to model the
class-conditional likelihood for this task.
Specifically, assuming each random variable (or pixel) to
be independent,
a {\sc cnn} models the likelihood function of the form
$P({\bf y}| I; \theta) = \prod_{m=1}^n p(y_m | I; \theta)$,
where $p(y_m = l| I; \theta)$ is the \textit{softmax}
probability (or the marginal) of assigning label $l$
to the $m$-th pixel, obtained by applying the
\textit{softmax}%
\footnote{The \textit{softmax} function is defined as
$\sigma(f_k) = \frac{e^{f_k}}{\sum_{j=0}^c e^{f_j}}$}
function to the {\sc cnn} outputs $f(y_m| I; \theta)$
such that
$p(y_m = l| I; \theta) \propto \exp(f(y_m = l| I; \theta))$.
Given a training dataset
$\mathcal{S} = \{ I_i, {\bf y}_i\}_{i=1}^N$, where $I_i$
and ${\bf y}_i$ represent the $i$-th image and its
corresponding ground-truth semantic segmentation,
respectively, the log-likelihood is maximized by
minimizing the cross-entropy loss function using the
back-propagation algorithm to obtain the optimal $\theta$.
At test time, for a given image, the learned $\theta$ is
used to obtain the \textit{softmax} probabilities for
each pixel.
These probabilities are either post-processed or used
directly to assign semantic labels to each pixel.

\section{Weakly-Supervised Semantic Segmentation}
As mentioned in \secref{sec:semSeg},
to find the optimal $\theta$ for the
semantic segmentation task,
we need a dataset with ground-truth pixel-level
semantic labels,
obtaining which is highly time-consuming and expensive:
for a given image,
annotating it with $20$ object classes takes nearly $20$ seconds,
while pixel-wise segmentation takes nearly $239.7$
seconds~\cite{Bearman16Point}.
This is highly non-scalable to higher numbers of images
and classes.
Motivated by this, we use an Expectation-Maximization (EM)
\cite{Dempster77EM,McLachlan97EMBook} based approach for
weakly-supervised semantic segmentation using only
image labels.
Let us denote $Z = \mathcal{L}\setminus l_0$
as the set of object labels we are interested in, and
a weak dataset as
$\mathcal{D} = \{ I_i, {\bf z}_i\}_{i=1}^N$
where $I_i$ and ${\bf z}_i \subseteq Z$ are the $i$-th
image and the labels present in the image.
The task is to learn an optimal $\theta$ using
$\mathcal{D}$.

\begin{figure}[t]
    \centering
    \includegraphics[width=.4\linewidth]{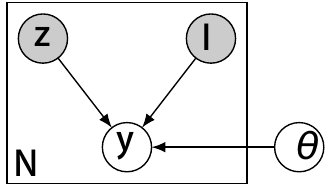}
    \caption{The graphical model. $I$ is the image and ${\bf z}$ is the set of labels present in the image. ${\bf y}$ is the latent variable (semantic segmentation) and $\theta$ is the set of parameters.}
     \label{fig:pgm}
\end{figure}

\subsection{The EM Algorithm}
\label{sec:em}
Similar to~\cite{Papandreou15EM},
we treat the unknown semantic segmentation
${\bf y}$ as the latent variable. Our probabilistic graphical model is of the following form
(\figref{fig:pgm}):
\begin{eqnarray}\label{eq:pgm}
P(I, {\bf y}, {\bf z}; \theta) = P(I)P({\bf y}| I, {\bf z}; \theta)P({\bf z}),
\end{eqnarray}
where we assume that
$P({\bf y}| I, {\bf z}; \theta) =
P({\bf y}| I; \theta)P({\bf y}| {\bf z})$.
Briefly, to learn $\theta$ while maximizing the above
joint probability distribution,
the three major steps of an EM algorithm  are:
(i) initialize the parameter $\theta_t$;
(ii) E-step: compute the expected complete-data
log-likelihood $F(\theta; \theta_t)$; and
(iii) M-step: update $\theta$ by maximizing
$F(\theta; \theta_t)$.
In what follows, we first talk about how to obtain a good initialization $\theta_t$ in order to avoid poor local maxima and then talk about optimizing parameters (E and M steps) for a given $\theta_t$.
\subsection{Initialization: Skipping Poor Local Maxima Using Bottom-Up Top-Down Cues}
\label{sec:initialModel}
It is well known that in case that the log-likelihood has several maxima or saddle points, an EM-based approach is highly susceptible to mediocre local maxima and a good initialization is crucial~\cite{JeffWu83EMConvergence}.
We argue that instead of initializing the algorithm with parameters learned for the classification task using the \textit{ImageNet} classification dataset~\cite{Deng09ImageNet}, as is done by most state-of-the-art methods irrespective of their nature, it is much more effective and intuitive to initialize with parameters learned for solving the task at hand - that is {\em semantic segmentation} - using the same dataset. This would allow for the full power of the dataset to be harnessed. For weakly-supervised semantic segmentation, however, the challenge is that only image-level labels are accessible during the training process. In the following we address this issue to obtain a good initialization $\theta_t$ by training an initial {\sc cnn} model over simple \textit{ImageNet} images for the weakly-supervised semantic segmentation task.


Let us denote $\mathcal{D}(I)$ as a subset of images from the \textit{ImageNet} dataset containing objects of the categories in which we are interested (for details of this dataset, see Section~\ref{sec:experiments}). Dataset $\mathcal{D}(I)$ mostly contains images with centered and clutter-free single objects, unlike the challenging PASCAL VOC 2012~\cite{Everingham14VOC12}. Given $\mathcal{D}(I)$, in order to train the initial model to obtain $\theta_t$, we need pixel-level semantic labels which are not available in the weakly-supervised setting. To circumvent this, we use the combination of a class-agnostic \textit{saliency} map~\cite{ChengPAMI,Hou2016DeepSaliency} (bottom-up cue) and a class-specific \textit{attention} map~\cite{Zhang16Attention} (top-down cue) to obtain an approximate ground-truth probability distribution of labels for each pixel. Intuitively, a saliency map gives the probability of each pixel belonging to any foreground class, and an attention map provides the probability of it belonging to the given object class. Combining these two maps allow us to obtain a very accurate ground-truth probability distribution of the labels for each pixel in the image (see Figure~\ref{fig:pixelMining}). 

Precisely, as shown in Algorithm~\ref{algo:miningPixels}, for a given simple image $I \in \mathcal{D}(I)$ and its corresponding class label $z \in Z$, we combine the attention and the saliency values per pixel to obtain $M(m) \in [0,1]$ for all the pixels in the image. The value of $M(m)$ for the $m$-th pixel denotes the probability of it being the $z$-th object category. Similarly, $1-M(m)$ is the probability of it being the background. The combination function $h(.,.)$ in Algorithm~\ref{algo:miningPixels} is a user-defined function that combines the saliency and the attention maps. In this work we employ the $\max$ function which takes the union of the attention and saliency maps (Figure~\ref{fig:pixelMining}), thus complement each other. Let us define the approximate ground-truth label distribution for the $m$-th pixel as $\delta_m^I$. Thus, $\delta_m^I \in [0,1]^{|\mathcal{L}|}$, where $\delta_m^I(z) = M(m)$ at the $z$-th index for the object category $z$, $\delta_m^I(0) = 1-M(m)$ at the $0$-th index for the background, and zero otherwise. Given $\delta_m^I$ for each pixel, we find $\theta_t$ by using a {\sc cnn} and optimizing the per-pixel cross-entropy loss $\sum_{k \in \mathcal{L}} \delta_m^I(k) \log p(k| I; \theta)$ between $\delta_m^I$ and $p(y_m| I, \theta)$, where $p(y_m | I, \theta)$ is the {\sc cnn} likelihood.
\begin{algorithm}[tb]
\caption{Approximate Ground Truth Distribution}
\label{algo:miningPixels}
\begin{algorithmic}[1]
\INPUT  Image $I$ with one object category; Image label ${z}$
\STATE $M = zeros (n)$, $n$ is the number of pixels.
\STATE ${\bf s} \leftarrow SaliencyMap(I)$~\cite{Hou2016DeepSaliency}
\STATE ${\bf a} \leftarrow AttentionMap(I, z)$~\cite{Zhang16Attention}
\FOR {each pixel $m \in I$}
\STATE $M(m) = h ({\bf s}(m), {\bf a}(m))$ 
\ENDFOR
\OUTPUT $M$
\end{algorithmic}
\end{algorithm}
Notice that using this approach to obtain the approximate ground-truth label distribution requires only image-level labels. No human annotator involvement is required. By using the probability value $M(m)$ directly instead of using a standard Dirac delta distribution makes the approach more robust to noisy attention and saliency maps. This approach can be seen as a way of mining class-specific noise-free pixels, and is motivated by the work of Bearman et al.~\cite{Bearman16Point} where humans annotate points and squiggles in complex images. Their work showed that instead of training a network with a fully-supervised dataset, the learning process can be sufficiently guided using only a few supervised pixels which are easy to obtain. Their approach still requires human intervention in order to obtain points and squiggles, whereas, our approach requires only image-level labels which makes it highly scalable.




\subsection{Optimizing Parameters}\label{sec:optimization}
Let us now talk about how to define and optimize the expected complete-data log-likelihood $F(\theta; \theta_t) $. By definition,
$F(\theta; \theta_t) = \sum_{\bf y} P({\bf y}| I,
{\bf z}; \theta_t) \log P(I, {\bf y}, {\bf z}; \theta)$, where the expectation is taken over the posterior over
the latent variables at a given set of parameters $\theta_t$,
denoted as $P({\bf y}| I, {\bf z}; \theta_t)$.
In the case of semantic segmentation, the latent space
is exponentially large $|{\mathcal{L}}|^n$,
therefore, computing $F(\theta; \theta_t)$ is infeasible.
However, as will be shown,
the independence assumption over the random variables,
namely
$P({\bf y}| I; \theta) = \prod_{m=1}^n p(y_m | I; \theta)$,
allows us to maximize $F(\theta; \theta_t)$ efficiently
by decomposition.
By using \equref{eq:pgm},
the independence assumption,
the identity
$\sum_{\bf y} P({\bf y}| I, {\bf z}; \theta_t) = 1$,
and ignoring the terms independent of
$\theta$, $F(\theta; \theta_t)$ can be written in a
simplified form as:
\begin{eqnarray}\label{eq:Estep}
\bar{F}(\theta; \theta_t) = \sum_{m=1}^n \sum_{\bf y} P({\bf y}| I, {\bf z}; \theta_t) \log p(y_m| I; \theta)
\end{eqnarray}
Without loss of generality, we can write
$P({\bf y}| I, {\bf z}; \theta_t) =
P({\bf y} \setminus y_m | I, {\bf z}, y_m; \theta_t)
p(y_m| I, {\bf z}; \theta_t)$,
and using the identity
$\sum_{{\bf y} \setminus y_m} P({\bf y}
\setminus y_m | I, {\bf z}, y_m; \theta_t) = 1$,
we obtain:
\begin{eqnarray}\label{eq:EstepFinal}
\bar{F}(\theta; \theta_t) = \sum_{m=1}^n \sum_{y_m \in \mathcal{L}} p(y_m| I, {\bf z}; \theta_t) \log p(y_m| I; \theta)
\end{eqnarray}
Therefore, the M-step parameter update,
which is maximizing $\bar{F}(\theta; \theta_t)$ w.r.t.
$\theta$, can be written as:
\begin{eqnarray}\label{eq:Mstep}
\theta_{t+1} = \argmax_{\theta} \sum_{m=1}^n \sum_{y_m \in \mathcal{L}} p(y_m| I, {\bf z}; \theta_t) \log p(y_m| I; \theta)
\end{eqnarray}
As mentioned in \secref{sec:em}, the latent posterior distribution is
$P({\bf y}| I, {\bf z}; \theta_t) =  P({\bf y}| I;
\theta_t) P({\bf y} | {\bf z})$,
where $P({\bf y}| I; \theta_t)$ is the likelihood obtained using the {\sc cnn} at a given $\theta_t$. The distribution $P({\bf y} | {\bf z})$ can be used to {\em regularize} the {\sc cnn} likelihood $P({\bf y}| I; \theta_t)$ by imposing constraints based on the image label information given in the dataset. Note that, $P({\bf y} | {\bf z})$ is independent of $\theta$ and is a task-specific user-defined distribution that depends on the image labels or object categories. For example, if we know that there are only two classes in a given training image such as `cat' and `person' out of many other possible classes, then we would like to push the latent posterior probability $P({\bf y}| I, {\bf z}; \theta_t)$ of absent classes to zero and increase the probability of the present classes. To impose this constraint, let us assume that similar to the likelihood, $P({\bf y} | {\bf z})$ also decomposes over the pixels and belongs to the exponential family distribution $p(y_m | {\bf z}) \propto \exp (g (y_m, {\bf z}))$, where $g(.,.)$ is the user-define function designed to impose the desired constraints. Thus, the posterior can be written as $p(y_m| I, {\bf z}; \theta_t) \propto \exp (f(y_m| I; \theta_t) + g (y_m, {\bf z}))$. In order to impose the above mentioned constraints, we use the following form of $g(.,.)$:
\begin{equation}
g(y_m, {\bf z}) =
\Bigg\{
\begin{array}[1]{c}
-\infty, \; \;if \; y_m \notin {\bf z} \cup l_0, \\
0,\; otherwise.
\end{array}
\label{eq:classPriorG}
\end{equation}
Practically speaking, imposing the above constraint is equivalent to obtaining \textit{softmax} probabilities for only those classes (including background) present in the image and assigning a probability of zero to all other classes. In other words, the above definition of $g(.,.)$ inherently defines $p(y_m| {\bf z})$ such that it is uniform for the classes present in the image including the background ($l_0$) and zero for the remaining ones. Other forms of $g(.,.)$ can also be used to impose different  task-specific label-dependent constraints.
\vspace{-4mm}
\paragraph{Cross entropy functions} Consider the parameter update (M-step) as defined in Eq.~\ref{eq:Mstep}. Solving this is equivalent to minimizing the cross entropy or the {\sc kl} divergence between the latent posterior distribution and the {\sc cnn} likelihood. Notice that, as opposed to~\cite{Papandreou15EM}, which uses a Dirac delta approximation $\hat{p}$ of the posterior distribution, where $\hat{p} (\hat{l}_m) = 1$ at $\hat{l}_m = \argmax_{l \in \mathcal{L}} p(y_m = l| I, {\bf z}; \theta_t)$ and otherwise zero, we use the true posterior distribution (or the regularized likelihood) itself. We argue that using a Dirac delta distribution imposes a {\em hard} constraint that is suitable only when we are very confident about the true label assignment (for example, in the fully-supervised setting where the label is provided by a human annotator). In the weakly-supervised setting where the latent posterior, which decides the label, can be noisy (mostly seen in the case of confusing classes), it is more suitable to use the true posterior distribution, obtained using the combination of the {\sc cnn} likelihood $p( y_m | I, \theta_t)$ and the class label-based prior $p(y_m| {\bf z})$. We propose to optimize $\theta$ by combining this true posterior distribution and its Dirac delta  approximation:
\begin{eqnarray}
\label{eq:entropyEquationFinal}
J_m(I, {\bf z}, \theta_t; \theta) = \sum_{y_m \in \mathcal{L}}  \bar{p}(y_m|I, {\bf z}; \theta_t) \log p(y_m| I; \theta)
\end{eqnarray}
where, $\bar{p}(y_m|I, {\bf z}; \theta_t) = (1- \epsilon)p(y_m|I, {\bf z}; \theta_t) + \epsilon \hat{p}(y_m)$. To investigate the interplay between the two terms, we define a criterion which we call the \textit{Relative Heuristic} to compute the value of $\epsilon$ given the pixel-wise latent posterior $p(y_m = l | I, {\bf z}; \theta)$ and a user-defined hyper-parameter $\eta \in [0, 1]$:
\begin{equation}
\epsilon =
\Bigg\{
\begin{array}[1]{c}
1, \; \;if \; r \geq \eta, \\
r, \; otherwise.
\end{array}
\label{eq:heuristic2}
\end{equation}
where $r = (p_1 - p_2)/p_1$, and $p_1$ and $p_2$ are the highest and the second highest probability values in the latent posterior distribution. Intuitively, $\eta = 0.05$ implies that the most probable score should be at least $5\%$ better compared to the second most probable score. 
\vspace{-4mm}
\paragraph{The IoU gain function} Along with minimizing the cross entropy losses as shown in the Eq.~\ref{eq:entropyEquationFinal}, in order to obtain better parameter estimate, we also maximize the probabilistic approximation of the intersection-over-union (IoU) between the posterior distribution and the likelihood~\cite{Ahmed15ExpectedIoU,Cogswell14IoUCNN,Nowozin14IoU}:
\begin{eqnarray}
\label{eq:iou}
\mathcal{J}_{IOU} (P({\bf y}|I, {\bf z}; \theta_t), P({\bf y}|I; \theta) ) \approx \nonumber \\
\frac{1}{|\mathcal{L}|} \sum_{l \in \mathcal{L}} \frac{ \sum_{m=1}^n  p^t_m(l) p^{\theta}_m(l) }{ {\sum_{m=1}^n} \{p^t_m(l) + p^{\theta}_m(l) - p^t_m(l) p^{\theta}_m(l) \}}
\end{eqnarray}
where, $p^t_m (l) = p(y_m = l| I, {\bf z}; \theta_t)$ and $p^{\theta}_m (l) = p(y_m = l| I; \theta)$. Refer to~\cite{Cogswell14IoUCNN} for further details about Eq.~\ref{eq:iou}. 


\paragraph{Overall objective function and the algorithm} Combining the cross entropy loss function (Eq.~\ref{eq:entropyEquationFinal}) and the IoU gain function (Eq.~\ref{eq:iou}), the M-step parameter update problem is:
\begin{eqnarray}
\label{eq:MstepFinal}
\theta_{t+1} = \argmax_{\theta} \sum_{m=1}^n J_m(I, {\bf z}, \theta_t; \theta) + \mathcal{J}_{IOU}
\end{eqnarray}
We use a {\sc cnn} model along with the back-propagation algorithm to optimize the above objective function. Recall that our evaluation is based on the PASCAL VOC 2012 dataset, therefore, during the M-step of the algorithm we use both the \textit{ImageNet} $\mathcal{D}(I)$ and the PASCAL trainval $\mathcal{D}(P)$ datasets (see Section~\ref{sec:experiments} for details). Our overall approach is summarized in Algorithm~\ref{algo:emFinal}.

\begin{algorithm}[tb]
\caption{Our Final Algorithm}
\label{algo:emFinal}
\begin{algorithmic}[1]
\INPUT Datasets $\mathcal{D}(P)$ and $\mathcal{D}(I)$; $\theta_0$; $\eta$; $K$
\STATE Use $\mathcal{D}(I)$ and $\theta_0$ to obtain initialization parameter $\theta_t$ using method explained in Section~\ref{sec:initialModel}.
\FOR {$k = 1 : K$}
\STATE $\theta \leftarrow \theta_t$
\FOR {each pixel $m$ in $\mathcal{D}(P) \cup \mathcal{D}(I)$}
\STATE Obtain latent posterior: $p_m(y_m| I, {\bf z}; \theta) \propto \exp (f_m(y_m| I; \theta) + g (y_m, {\bf z}))$
\ENDFOR
\STATE Optimize Eq.~\ref{eq:MstepFinal} using {\sc cnn} to update $\theta_t$. 
\ENDFOR
\end{algorithmic}
\end{algorithm}

\begin{table*}
	\centering
	\scalebox{0.88}{
		\begin{tabular*}{\textwidth}{cccccct{1.3cm}t{1.3cm}}
			\toprule[1.2pt]
			\multicolumn{2}{c}{{\bf Method}} & {\bf Dataset} & {\bf Dependencies} & {\bf Supervision} & {\bf CRF}~\cite{Krahenbuhl11FullyCRF} & {\bf mIoU (Val)}  & {\bf mIoU (Test)} \\
			\cline{1-8}
			\multicolumn{2}{c}{\multirow{2}*{EM Adapt~\cite{Papandreou15EM}}} & \multirow{2}*{$\mathcal{D}(I) , \mathcal{D}(P)$} & \multirow{2}*{No} & \multirow{2}*{Image labels} & \xmark & $-$  & $-$ \\
			\cline{6-8}
			& & & & & \cmark & $38.2\%$ & $39.6\%$
			\\
			\cline{1-8}
			\multicolumn{2}{c}{\multirow{4}*{{\sc ccnn}~\cite{Pathak15CCNN}}} & \multirow{4}*{$\mathcal{D}(I) , \mathcal{D}(P)$} & \multirow{2}*{No} & \multirow{4}*{Image labels} & \xmark & $33.3\%$  & $35.6\%$ \\
			\cline{6-8}
			& & & & & \cmark & $35.3\%$ & $-$ \\
			\cline{4-4} \cline{6-8}
			& & & \multirow{2}*{Class size} &  & \xmark & $\underline{\emph{40.5}}\%$ & $\underline{\emph{43.3}}\%$ \\
			\cline{6-8}
			& & & & & \cmark & $42.4\%$ & $45.1\%$ \\
			\cline{1-8}
			\multicolumn{2}{c}{\multirow{2}*{{\sc sec}~\cite{Kolesnikov16SEC}}} & \multirow{2}*{$\mathcal{D}(I) , \mathcal{D}(P)$} & Saliency~\cite{Simonyan14Saliency} \& & \multirow{2}*{Image labels} & \xmark & $\underline{\emph{44.3}}\%$  & $-$ \\
			\cline{6-8}
			& & & Localization~\cite{Zhou16Localization} & & \cmark &  $50.7\%$ & $51.7\%$
			\\
			\cline{1-8}
			\multicolumn{2}{c}{\multirow{3}*{{\sc mil}~\cite{Pinheiro15Image2Pixel}}} & \multirow{3}*{$\mathcal{D}(I)$}& Superpixels~\cite{Felzenszwalb04Superpixel} &\multirow{3}*{Image labels} & \multirow{3}*{\xmark} & $36.6\%$ & $35.8\%$
			\\ \cline{4-4} \cline{7-8}
			& & & BBox BING~\cite{Cheng14Bing} & & & $\underline{\emph{37.8}}\%$ & $\underline{\emph{37.0}}\%$
			\\ \cline{4-4} \cline{7-8}
			& & & MCG~\cite{Arbelaez14MCG} & & & $42.0\%$ & $40.6\%$
			\\
			\cline{1-8}
			\multicolumn{2}{c}{\multirow{5}*{{\sc wtp}~\cite{Bearman16Point}}}  & \multirow{5}*{$\mathcal{D}(I) , \mathcal{D}(P)$} & \multirow{5}*{Objectness~\cite{Alexe12Objectness}} & Image labels &  \multirow{5}*{$-$} & $\underline{\emph{32.2}}\%$ & $-$
			\\ \cline{5-5} \cline{7-8}
			& & & & Image labels + &  & \multirow{2}*{$42.7\%$} & \multirow{2}*{$-$}
			\\
			& & & & 1 Point/Class &  & &
			\\ \cline{5-5} \cline{7-8}
			& & & & Image labels + &  & \multirow{2}*{$49.1\%$} & \multirow{2}*{$-$}
			\\
			& & & & 1 Squiggle/Class &  & &
			\\
			\cline{1-8}
			\multicolumn{2}{c}{{\sc stc}~\cite{Yunchao15STC}} & $\mathcal{D}(I), \mathcal{D}(P), \mathcal{D}(F)$ & Saliency~\cite{Jiang13Saliency} & Image labels & \cmark & $49.8\%$ & $51.2\%$
			\\
			\cline{1-8}
			\multicolumn{2}{c}{\multirow{4}*{AugFeed~\cite{Qi16AugFeedBack}}} &  \multirow{4}*{$\mathcal{D}(I) , \mathcal{D}(P)$} &  \multirow{2}*{SS~\cite{Uijlings13SelectiveSearch}} &  \multirow{4}*{Image labels} & \xmark & $\underline{\emph{46.98}}\%$  & $\underline{\emph{47.8}}\%$ \\
			\cline{6-8}
			& & & & &\cmark & $52.62\%$ & $52.7\%$ \\
			\cline{4-4} \cline{6-8}
			& & & \multirow{2}*{MCG~\cite{Arbelaez14MCG}}  & & \xmark & $50.41\%$ & $50.6\%$ \\
			\cline{6-8}
			& & & & & \cmark & $54.34\%$ & $55.5\%$ \\
			\toprule[1.2pt]
			\multirow{4}*{\bf Ours}  & \multirow{2}*{Initial Model} & \multirow{2}*{$\mathcal{D}(I)$} & & \multirow{4}*{Image labels} & \xmark &  $53.53\%$  & $54.34\%$ \\
			\cline{6-8}
            & & & Saliency~\cite{Hou2016DeepSaliency} \& & & \cmark & $ 55.19\%$  & $56.24\%$ \\
            \cline{2-2} \cline{6-8}
			& \multirow{2}*{Final} & \multirow{2}*{$\mathcal{D}(I) , \mathcal{D}(P)$}& Attention~\cite{Zhang16Attention} & & \xmark & ${\bf 56.91}\%$  & ${\bf 57.74}\%$ \\
            \cline{6-8}
            & & & & & \cmark & $58.71\%$  & $59.58\%$ \\
			\cline{1-8}
			\toprule[1.2pt]
		\end{tabular*}
	}
	\caption{\small \linespread{0.7} {\bf Comparison table}. All dependencies, datasets, and degrees of supervision used by most of the existing methods for weakly-supervised semantic segmentation. Table~\ref{tab:dependencyComparison} complements this table by providing the degrees of supervision used by the dependencies themselves. $\mathcal{D}(I)$ is the simple filtered \textit{ImageNet} dataset, $\mathcal{D}(P)$ is the complex filtered PASCAL VOC 2012 dataset (see Section~\ref{sec:experiments}) and $\mathcal{D}(F)$ contains $41$K images from {\em Flickr} \cite{Yunchao15STC}. Note that the cross validation of CRF hyper-parameters and the training of {\sc mcg} are performed using a fully-supervised pixel-level semantic segmentation dataset. To be fair, our method should only be compared with the numbers shown in italic with underline.}
	\label{tab:methodComparision}
\end{table*}


\section{Experimental Results and Comparisons}
\label{sec:experiments}
We show the efficacy of our method on the challenging PASCAL VOC 2012 benchmark and outperform all existing state-of-the-art methods by a large margin. Specifically, we improve on the current state-of-the-art weakly-supervised method using only image labels\cite{Qi16AugFeedBack} by 10\%.

\subsection{Setup}

\paragraph{Dataset $\mathcal{D}(I)$.} To train our initial model (Section~\ref{sec:initialModel}), we download $80,000$ images from the {\em ImageNet} dataset which contain objects in the $20$ foreground object categories of the PASCAL VOC 2012 segmentation task. We filter this dataset using simple heuristics. First, we discard images with width or height less than 200 or greater than 500 pixels. Using the attention model of \cite{Zhang16Attention} (which is trained with only image-level labels), we generate an attention map for each image and record the most probable class label with its corresponding probability. We discard the image if (i) its most probable label does not match the image label or (ii) its most probable label matches the image label but its corresponding probability is less than 0.2. We then generate saliency maps using the saliency model of~\cite{Hou2016DeepSaliency} (which is trained with only class-agnostic saliency masks). For the remaining images, we combine attention and saliency by finding the pixel-wise intersection between the saliency and the attention binary masks. The saliency mask is obtained by setting the pixel's value to 1 if its corresponding saliency probability is greater than 0.5. The same is done to obtain the attention mask. For each object category, the images are sorted by this intersection area (i.e. the number of overlapping pixels between the two masks) with the intuition that larger intersections correspond to higher quality saliency and attention maps. The top 1500 images are then selected for each category, with the exception of the `person' category where the top 2500 images are kept, and any category with fewer than 1500 images, in which case all images are kept. {\em This complete filtering process leaves us with $24,000$ simple images}, which contain uncluttered and mainly-centered single objects. We denote this dataset as $\mathcal{D}(I)$ and will make this dataset publicly available. {\em We highlight that $\mathcal{D}(I)$ does not contain any additional images relative to those used by other weakly supervised works} (see Dataset column in Table~\ref{tab:methodComparision}).

\paragraph{Datasets $\mathcal{D}(P)$ and $\mathcal{D}(I)$ for M-step.} For the M-step, we use a filtered subset of PASCAL VOC 2012 images, denoted $\mathcal{D}(P)$, and a subset of $\mathcal{D}(I)$. To obtain $\mathcal{D}(P)$, we take complex PASCAL VOC 2012 images ($10,582$ in total, made up of $1,464$ training images \cite{Everingham14VOC12} and the extra images provided by~\cite{Hariharan11trainAug}), and use the trained initial model (i.e. $\theta_t$) to generate a (hard) ground-truth segmentation for each. The hard segmentations are obtained by assigning each pixel with the class label that has the highest probability. The ratio of the foreground area to the whole image area (where area is the sum of the number of pixels) is computed, and if the ratio is below 0.05, the image is discarded. This leaves $10,000$ images. We also further filter $\mathcal{D}(I)$: using the trained initial model, we generate (hard) segmentations for all simple \textit{ImageNet} images in $\mathcal{D}(I)$. We compute the intersection area (as above) between the attention mask and the predicted segmentation (rather than the saliency mask as before) and select the top $10,000$ of $24,000$ images based on this metric. Together $\mathcal{D}(P)$ and this subset of $\mathcal{D}(I)$ consist of $20,000$ images used for the M-step.

\paragraph{CNN architecture and parameter settings} Similar to~\cite{Kolesnikov16SEC,Papandreou15EM,Yunchao15STC}, both our initial model and our EM model are based on the largeFOV DeepLab architecture~\cite{Chen15DeepLabV1}. We use simple bilinear interpolation to map the downsampled feature maps to the original image size as suggested in~\cite{Long15FCN}. We use the publicly available Caffe toolbox~\cite{Jia14Caffe} for our implementation. We use weight decay $(0.0005)$, momentum $(0.9)$, and iteration size $(10)$ for gradient accumulation. The learning rate is $0.001$ at the beginning and is divided by $10$ every $10$ epochs. We use a batch size of $1$ and randomly crop the input image to $321 \times 321$. Images with width or height less than $321$ are padded with the mean pixel values and the corresponding places in the ground-truth are padded with ignore labels to nullify the effect of padding. We flip the images horizontally, resulting in an augmented set twice the original one. We train our networks for $30$K iterations by optimizing Eq.~\ref{eq:MstepFinal} as per Algorithm~\ref{algo:emFinal} with $\eta = 0.05$ and $K=2$. Performance gains beyond two EM iterations were not significant compared to the computational cost.
\begin{table*}
	\centering
	\setlength\tabcolsep{2pt}
	\footnotesize
	\scalebox{0.95}{
	\begin{tabular*}{\textwidth}{c|c|c|ccccccccccccccccccccc|c} \cline{1-25}
		Data & Method & CRF & bkg & plane & bike & bird & boat & bottle & bus & car & cat & chair & cow & table & dog & horse & motor & person & plant & sheep & sofa & train & tv & mIoU \\ \cline{1-25}
		\multirow{4}*{Val}& \multirow{2}*{Initial} & \xmark & 87.1 & 74.7 & 29.0 & 69.8 & 55.8 & 55.6 & 73.3 & 65.2 & 63.4 & 15.8 & 61.5 & 15.9 & 60.0 & 56.4 & 57.5 & 53.7 & 32.9 & 65.6 & 23.9 & 64.6 & 42.2 & $53.53$ \\
		& & \cmark & 87.7  & 79.7 & 32.6 & 73.4  & 58.4 & 57.8 & 74.3 & 64.8 & 66.0  & 15.9  & 63.1 & 15.0  &  62.3 & 59.6 & 57.7 & 54.9 & 33.8 & 69.1 & 23.7 & 65.0 & 44.3  & $55.19$  \\
		& \multirow{2}*{Final} & \xmark & 87.8 & 72.4 & 28.7 & 67.9 & 58.8 & 55.8 & 78.0 & 69.7 & 70.2 & 17.8 & 63.3 & 23.2 & 65.7 & 60.5 & 63.1 & 58.7 & 40.0 & 68.2 & 28.9 & 70.9 & 45.5 & $\bf{56.91}$ \\
	    & & \cmark & 88.6  & 76.1 & 30.0 & 71.1 & 62.0 & 58.0 & 79.2 & 70.5 & 72.6 & 18.1 & 65.8 & 22.3 & 68.4 & 63.5 & 64.7 & 60.0 & 41.5 & 71.7 & 29.0 & 72.5 & 47.3 & $58.71$  \\ \cline{1-25}

		\multirow{4}*{Test}& \multirow{2}*{Initial} & \xmark & 87.9 & 69.2 & 29.2 & 74.9 & 41.7 & 53.4 & 70.6 & 69.6 & 59.9 & 18.3 & 66.1 & 24.9 & 62.5 & 63.3 & 68.8 & 55.4 & 33.7 & 63.8 & 18.6 & 64.3 & 44.9 & $54.35$ \\
		& & \cmark & 88.5  & 72.6 & 32.6  & 80.3  & 44.6 & 55.4  & 70.9 & 69.6  & 62.2 & 18.9 & 68.4 & 24.6  & 65.2 & 66.8 & 71.2 & 57.2 & 37.2 & 66.7 & 17.4 & 64.8 & 45.9  & $56.24$  \\
		& \multirow{2}*{Final} & \xmark &  88.2 & 69.5 & 29.7 & 72.2 & 45.1 & 57.3 & 73.2 & 72.7 & 69.3 & 20.5 & 65.4 & 33.5 & 67.8 & 64.0 & 72.3 & 58.9 & 45.5 & 69.8 & 26.8 & 63.8 & 46.8 & $\bf{57.74}$  \\
	    & & \cmark & 88.9  & 72.7  & 31.0 & 76.3  & 47.7 & 59.2 & 74.3 & 73.2  & 71.7 & 19.9 & 67.1 & 34.0 & 70.3 & 66.6 & 74.4 & 60.2 & 48.1 & 73.1 & 27.8 & 66.9 & 47.9 & $59.58$  \\ \cline{1-25}	
	\end{tabular*}
	}
	\vspace{2pt}
	\caption{\small Per-class results (in $\%$) on PASCAL VOC 2012 {\em val} and {\em test} sets. Initial shows results for initial model trained using $\mathcal{D}(I)$ for EM algorithm initialization. Final shows results at final EM iteration ($K=2$). Results show with and without application of post-process CRF~\cite{Krahenbuhl11FullyCRF} (although we do not consider using a CRF as true weak supervision, see Section~\ref{sec:experiments}).
	}\label{tab:perClassPerformance}
\end{table*}

\subsection{Results, Comparisons, and Analysis}
We provide Table~\ref{tab:methodComparision} and~\ref{tab:dependencyComparison} for an extenstive comparison between our and current methods, their dependencies, and degrees of supervision. Regarding the dependencies of our method, our {\em saliency} network~\cite{Hou2016DeepSaliency} is trained using salient region masks. These masks are class-agnostic, therefore, once trained the network can be used for any semantic object category, so there is no issue with scalability and no need to train the {\em saliency} network again for new object categories. Our second dependency, the {\em attention} network~\cite{Zhang16Attention} is trained using solely image labels.
\begin{table}
	\centering
	\linespread{1.3}
	\scalebox{0.9}{
		\begin{tabular*}{\textwidth}{x{4cm}x{4cm}}
			\cline{1-2}
			{\bf Dependency} & {\bf Supervision} \\ \cline{1-2}
			\multirow{2}*{Class size}  & Image labels \\
			& + Bboxes \\ \cline{1-2}
			\multirow{3}*{Saliency} & ~\cite{Simonyan14Saliency} Image labels \\ \cline{2-2}
			& \cite{Jiang13Saliency} Bboxes \\ \cline{2-2}
			& ~\cite{Hou2016DeepSaliency} Saliency masks \\ \cline{2-2}
			\cline{1-2}
			Attention~\cite{Zhang16Attention} & Image labels  \\ \cline{1-2}
			\multirow{2}*{Objectness~\cite{Alexe12Objectness} }  & Image labels \\
			& + Bboxes \\ \cline{1-2}
			Localization~\cite{Zhou16Localization}  & Image labels \\ \cline{1-2}
			Superpixels~\cite{Felzenszwalb04Superpixel} &  None \\ \cline{1-2}
			Bbox BING~\cite{Cheng14Bing} & Bboxes \\ \cline{1-2}
			MCG~\cite{Arbelaez14MCG} & Pixel labels \\ \cline{1-2}
			SS~\cite{Uijlings13SelectiveSearch} & Bboxes \\ \cline{1-2}
			\multirow{2}*{CRF~\cite{Krahenbuhl11FullyCRF}} & Pixel labels \\
			& (parameter cross-val) \\ \cline{1-2}
		\end{tabular*}}
		\caption{\small \linespread{0.7} {\bf Dependency table}.
			The degrees of supervision required by the
			dependencies of several weakly-supervised methods
			for the semantic segmentation task.
			`Bboxes' represents bounding boxes.
		}\label{tab:dependencyComparison}
	\end{table}

\paragraph{State-of-the-art.} Our method outperforms all existing state-of-the-art methods by a very high margin. The most directly comparable method in terms of supervision and dependencies is {\em AugFeed}~\cite{Qi16AugFeedBack} which uses super-pixels. Our method obtains almost $10\%$ better mIoU than {\em AugFeed} on both the \textit{val} and \textit{test} sets. Even if we disregard `equivalent' supervision and dependencies, our method is still almost $2.6\%$ and $2.2\%$ better than the best method in the \textit{val} and \textit{test} sets, respectively. Table~\ref{tab:perClassPerformance} shows class-wise performance of our method.

\paragraph{Simplicity vs sophistication.} The initial model is essential to the success of our method. We train this model in a very simple and intuitive way by employing a filtered subset of simple {\em ImageNet} images and training for the semantic segmentation task. Importantly, this process uses only image labels and is fully automatic, requiring no human intervention. The learned $\theta_t$ provides a {\em very} good initialization for the EM algorithm, enabling it to avoid poor local maxima. This is shown visually in Figure~\ref{fig:visualComp}: the initial model (third column) is already a good prediction, with the first and second EM iterations (fourth and fifth columns), improving the semantic segmentation even further. We highlight that with this simple approach, surprisingly, our initial model beats {\em all} current state-of-the-art methods, which are more complex and often use higher degrees of supervision. By implementing this intuitive modification, we believe that many methods can easily boost their performance.

\paragraph{To CRF or not to CRF?}
In our work, we specifically choose not to employ a CRF~\cite{Krahenbuhl11FullyCRF} as a post-processing step nor inside our models, during training and testing, for the following reasons: (1) CRF hyper-parameters are normally cross validated over a fully-supervised pixel-wise segmentation dataset which contradicts a fully ``weak" supervision. This is likewise the case for {\sc mcg}~\cite{Arbelaez14MCG} which is trained on a pixel-level semantic segmentation dataset. (2) The CRF hyper-parameters are incredibly sensitive, and if we wish to extend our framework to incorporate new object categories, this would require a pixel-level annotated dataset of the new categories along with the old ones for the cross-validation of the CRF hyper-parameters. This is highly non-scalable. For completeness, however, we include our method with a CRF applied (Table~\ref{tab:methodComparision} \&~\ref{tab:perClassPerformance}) which boosts our accuracy by $1.8$\%. We note that even without a CRF, our best approach exceeds the state-of-the-art (which uses a CRF {\em and} a higher degree of supervision) by $2.2$\% on the {\em test} set.

\section{Conclusions and Future Work}
We have addressed weakly-supervised semantic segmentation using only image labels. We proposed an EM-based approach and focus on the three key components of the algorithm: (i) initialization, (ii) E-step and (iii) M-step. Using only the image labels of a filtered subset of \textit{ImageNet}, we learn a set of parameters for the semantic segmentation task which provides an informed initialization of our EM algorithm. Following this, with each EM iteration, we empirically and qualitatively verify that our method improves the segmentation accuracy on the challenging PASCAL VOC 2012 benchmark. Furthermore, we show that our method outperforms all state-of-the-art methods by a large margin.

Future directions include making our method more robust to noisy labels, for example, when images downloaded from the Internet have incorrect labels, as well as better handling images with multiple classes of objects.
{\small
\bibliographystyle{ieee}
\bibliography{dokaniaBibliography}

\begin{thebibliography}{10}\itemsep=-1pt

\bibitem{Ahmed15ExpectedIoU}
F.~Ahmed, D.~Tarlow, and D.~Batra.
\newblock Optimizing expected intersection-over-union with
  candidate-constrained crfs.
\newblock In {\em ICCV}, 2015.

\bibitem{Alexe12Objectness}
B.~Alexe, T.~Deselares, and V.~Ferrari.
\newblock Measuring the objectness of image windows.
\newblock In {\em PAMI}, 2012.

\bibitem{Arbelaez14MCG}
P.~Arbelaez, J.~Pont-Tuset, J.~Barron, F.~Marques, and J.~Malik.
\newblock Multiscale combinatorial grouping.
\newblock In {\em CVPR}, 2014.

\bibitem{Bearman16Point}
A.~Bearman, O.~Russakovsky, V.~Ferrari, and F.-F. Li.
\newblock What's the point: Semantic segmentation with point supervision.
\newblock In {\em ECCV}, 2016.

\bibitem{Chandra16FastExactSemantic}
S.~Chandra and I.~Kokkinos.
\newblock Fast, exact and multi-scale inference for semantic image segmentation
  with deep gaussian crfs.
\newblock In {\em ECCV}, 2016.

\bibitem{Chen15DeepLabV1}
L.-G. Chen, G.~Papandreou, I.~Kokkinos, K.~Murphy, and A.~L. Yuille.
\newblock Semantic image segmentation with deep convolutional nets and fully
  connected.
\newblock In {\em ICLR}, 2015.

\bibitem{Cheng14Bing}
M.~Cheng, Z.~Zhang, W.~Lin, and P.~H.~S. Torr.
\newblock Bing: Binarized normed gradients for objectness estimation at 300fps.
\newblock In {\em CVPR}, 2014.

\bibitem{ChengPAMI}
M.-M. Cheng, N.~J. Mitra, X.~Huang, P.~H.~S. Torr, and S.-M. Hu.
\newblock Global contrast based salient region detection.
\newblock {\em IEEE TPAMI}, 37(3):569--582, 2015.

\bibitem{Cogswell14IoUCNN}
M.~Cogswell, X.~Lin, S.~Purushwalkam, and D.~Batra.
\newblock Combining the best of graphical models and convnets for semantic
  segmentation.
\newblock In {\em arXiv:1412.4313}, 2014.

\bibitem{Dempster77EM}
A.~P. Dempster, N.~M. Laird, and D.~B. Rubin.
\newblock Maximum likelihood from incomplete data via the em algorithm.
\newblock In {\em Journal of the Royal Statistical Society}, 1977.

\bibitem{Deng09ImageNet}
J.~Deng, W.~Dong, R.~Socher, L.-J. Li, K.~Li, and L.~Fei-Fei.
\newblock {ImageNet: A Large-Scale Hierarchical Image Database}.
\newblock In {\em CVPR}, 2009.

\bibitem{Everingham14VOC12}
M.~Everingham, S.~M.~A. Eslami, L.~V. Gool, C.~Williams, J.~Winn, and
  A.~Zisserman.
\newblock The pascal visual object classes challenge a retrospective.
\newblock In {\em IJCV}, 2014.

\bibitem{Felzenszwalb04Superpixel}
P.~F. Felzenszwalb and D.~P. Huttenlocher.
\newblock Efficient graph based image segmentation.
\newblock In {\em IJCV}, 2004.

\bibitem{Hariharan11trainAug}
B.~Hariharan, P.~Arbelaez, L.~Bourdev, S.~Maji, and J.~Malik.
\newblock Semantic contours from inverse detectors.
\newblock In {\em ICCV}, 2011.

\bibitem{Hou2016DeepSaliency}
Q.~Hou, M.-M. Cheng, X.-W. Hu, A.~Borji, Z.~Tu, and P.~Torr.
\newblock Deeply supervised salient object detection with short connections.
\newblock In {\em IEEE CVPR}, 2017.

\bibitem{JeffWu83EMConvergence}
C.~F. Jeff~Wu.
\newblock On the convergence properties of the em algorithm.
\newblock In {\em The Annals of Statistics}, 1983.

\bibitem{Jia14Caffe}
Y.~Jia, E.~Shelhamer, J.~Donahue, S.~Karayev, J.~Long, R.~Girshick,
  S.~Guadarrama, and T.~Darrell.
\newblock Caffe: Convolutional architecture for fast feature embedding.
\newblock In {\em ACM International Conference on Multimedia}, 2014.

\bibitem{Jiang13Saliency}
H.~Jiang, J.~Wang, Z.~Yuan, Y.~Wu, Z.~N., and S.~Li.
\newblock Salient object detection: A discriminative regional feature
  integration approach.
\newblock In {\em CVPR}, 2013.

\bibitem{Kolesnikov16SEC}
A.~Kolesnikov and C.~H. Lampert.
\newblock {Seed, Expand and Constrain: Three Principles for Weakly-Supervised
  Image Segmentation}.
\newblock In {\em ECCV}, 2016.

\bibitem{Krahenbuhl11FullyCRF}
P.~Krahenbuhl and V.~Koltun.
\newblock Efficient inference in fully connected crfs with gaussian edge
  potentials.
\newblock In {\em NIPS}, 2011.

\bibitem{Kullback51KLdivergence}
S.~Kullback and R.~A. Leibler.
\newblock On information and sufficiency.
\newblock {\em The Annals of Mathematical Statistics}, 1951.

\bibitem{Lin16ScribbleSup}
D.~Lin, J.~Dai, J.~Jia, K.~He, and J.~Sun.
\newblock Scribblesup: Scribble-supervised convolutional networks for semantic
  segmentation.
\newblock In {\em CVPR}, 2016.

\bibitem{Lin14COCO}
T.~Lin, M.~Maire, S.~Belongie, J.~Hays, P.~Perona, D.~Ramanan, P.~Doll{\'a}r,
  and C.~L. Zitnick.
\newblock Microsoft {COCO}: Common objects in context.
\newblock In {\em ECCV}, 2014.

\bibitem{Long15FCN}
J.~Long, E.~Shelhamer, and T.~Darrell.
\newblock Fully convolutional networks for semantic segmentation.
\newblock In {\em CVPR}, 2015.

\bibitem{McLachlan97EMBook}
G.~J. McLachlan and T.~Krishnan.
\newblock {\em {T}he {EM} algorithm and extensions}.
\newblock Wiley, 1997.

\bibitem{Nowozin14IoU}
S.~Nowozin.
\newblock Optimal decisions from probabilistic models: the
  intersection-over-union case.
\newblock In {\em CVPR}, 2014.

\bibitem{Papandreou15EM}
G.~Papandreou, L.-C. Chen, K.~P. Murphy, and A.~L. Yuille.
\newblock Weakly- and semi-supervised learning of a {DCNN} for semantic image
  segmentation.
\newblock In {\em ICCV}, 2015.

\bibitem{Pathak15CCNN}
D.~Pathak, P.~Krahenbuhl, and T.~Darrell.
\newblock Constrained convolutional neural networks for weakly supervised
  segmentation.
\newblock In {\em ICCV}, 2015.

\bibitem{Pathak14MIL}
D.~Pathak, E.~Shelhamer, J.~Long, and T.~Darrell.
\newblock Fully convolutional multi-class multiple instance learning.
\newblock In {\em ICLR}, 2014.

\bibitem{Pinheiro15Image2Pixel}
P.~O. Pinheiro and R.~Collobert.
\newblock From image-level to pixel-level labeling with convolutional networks.
\newblock In {\em CVPR}, 2015.

\bibitem{Qi16AugFeedBack}
X.~Qi, Z.~Liu, J.~Shi, H.~Zhao, and J.~Jia.
\newblock Augmented feedback in semantic segmentation under image level
  supervision.
\newblock In {\em ECCV}, 2016.

\bibitem{Simonyan14Saliency}
K.~Simonyan, A.~Vedaldi, and A.~Zisserman.
\newblock Deep inside convolutional networks: Visualising image classification
  models and saliency maps.
\newblock In {\em ICLR}, 2014.

\bibitem{Uijlings13SelectiveSearch}
J.~R.~R. Uijlings, K.~E.~A. van~de Sande, T.~Gevers, and A.~W.~M. Smeulders.
\newblock Selective search for object recognition.
\newblock In {\em IJCV}, 2013.

\bibitem{Xu15VariousWeak}
J.~Xu, A.~Schwing, and R.~Urtasun.
\newblock Learning to segment under various forms of weak supervision.
\newblock In {\em CVPR}, 2015.

\bibitem{Yunchao15STC}
W.~Yunchao, L.~Xiaodan, C.~Yunpeng, S.~Xiaohui, M.-M. Cheng, Z.~Yao, and
  Y.~Shuicheng.
\newblock {STC: A simple to complex framework for weakly-supervised semantic
  segmentation}.
\newblock In {\em arXiv:1509.03150}, 2015.

\bibitem{Zhang16Attention}
J.~Zhang, Z.~Lin, J.~Brandt, X.~Shen, and S.~Sclaroff.
\newblock Top-down neural attention by excitation backprop.
\newblock In {\em ECCV}, 2016.

\bibitem{Zheng15CRFRNN}
S.~Zheng, S.~Jayasumana, B.~Romera-Paredes, V.~Vineet, Z.~Su, D.~Du, C.~Huang,
  and P.~Torr.
\newblock Conditional random fields as recurrent neural networks.
\newblock In {\em ICCV}, 2015.

\bibitem{Zhou16Localization}
B.~Zhou, A.~Khosla, A.~Lapedriza, A.~Oliva, and A.~Torralba.
\newblock Learning deep features for discriminative localization.
\newblock In {\em CVPR}, 2016.

\end{thebibliography}
}

\newcommand{\addFig}[1]{\includegraphics[width=0.2\linewidth]{#1}}
\begin{figure*}[ht]
	\centering
	\scalebox{0.7}{
		\begin{tabular*}{\linewidth}{ccccc}
			\addFig{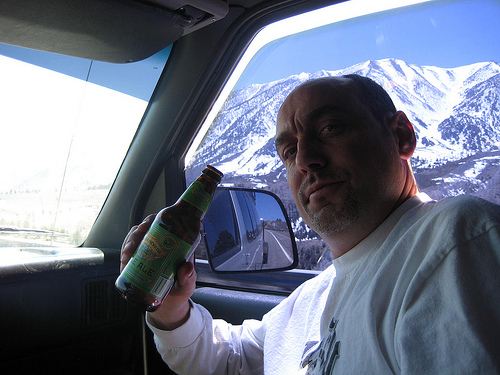} & 
			\addFig{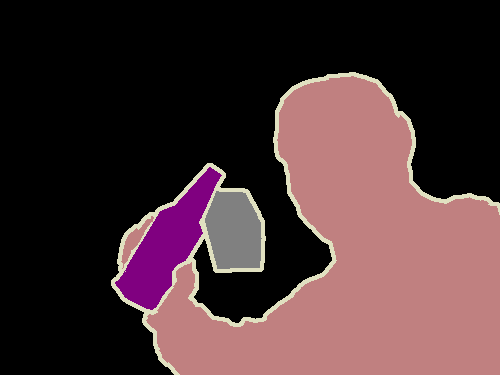} &
			\addFig{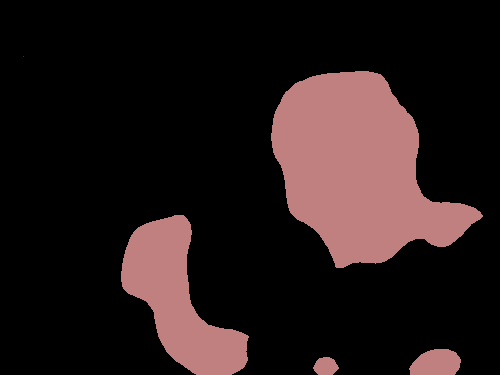} &
			\addFig{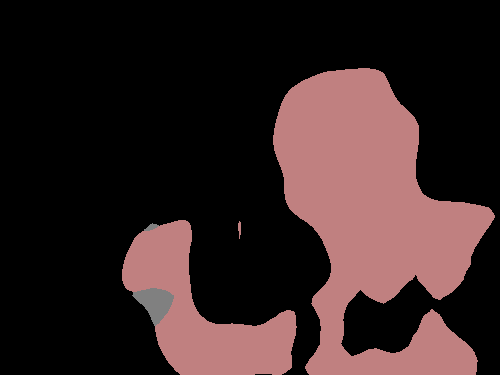} &
			\addFig{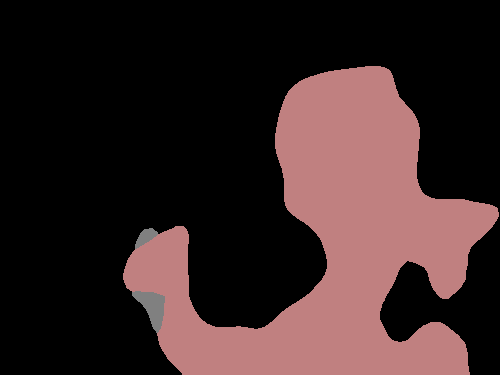} \\
			\addFig{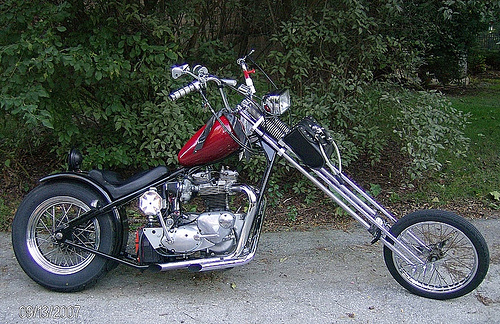} & 
			\addFig{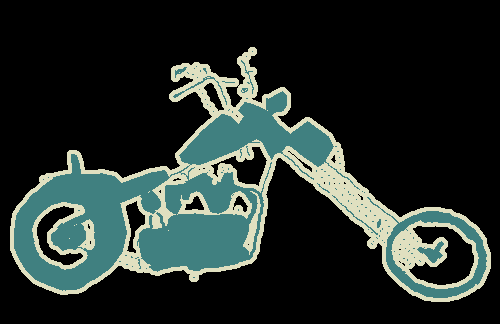} &
			\addFig{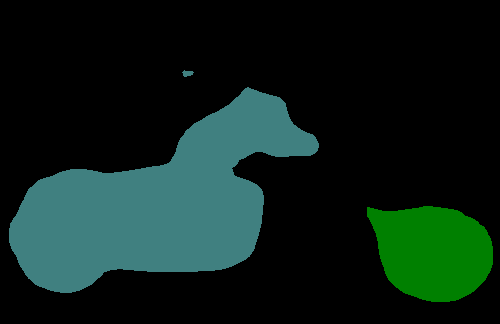} &
			\addFig{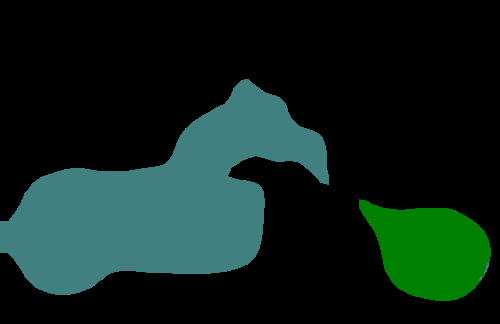} &
			\addFig{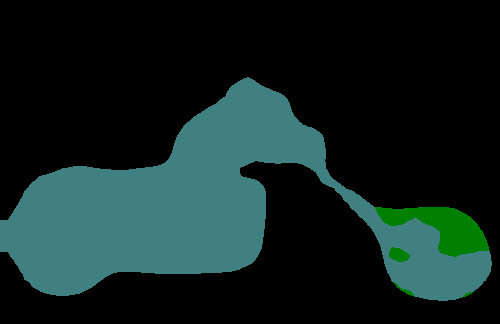} \\
			\addFig{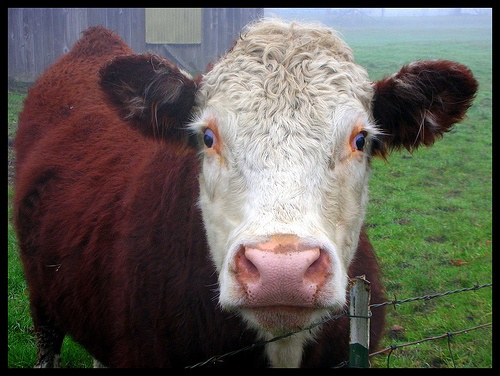} &
			\addFig{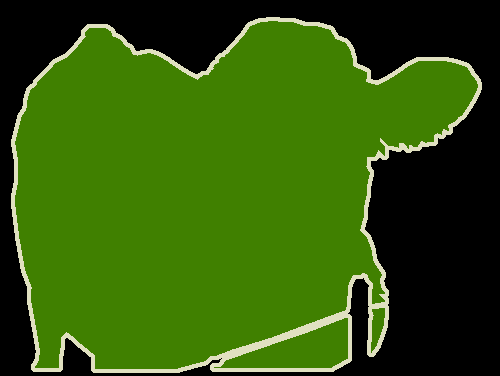} &
			\addFig{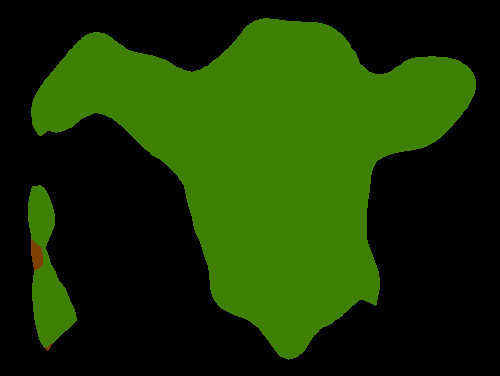} &
			\addFig{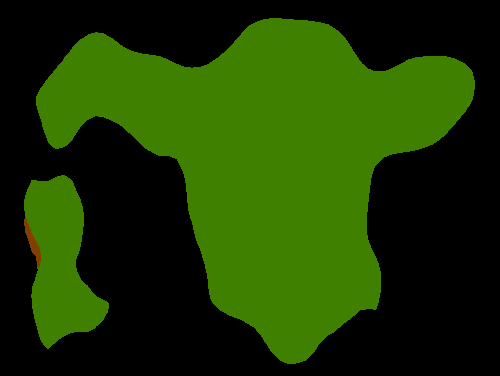} &
			\addFig{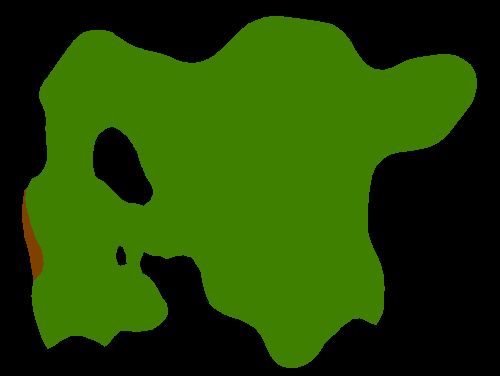} \\
			\addFig{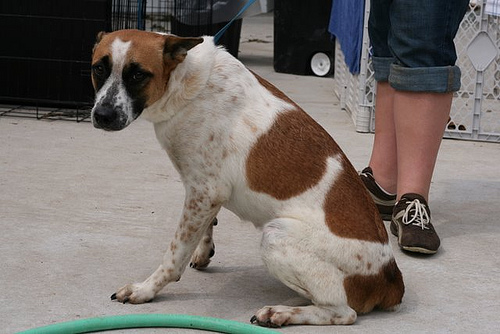} & 
			\addFig{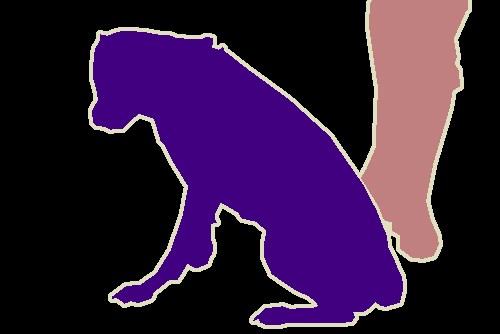} &
			\addFig{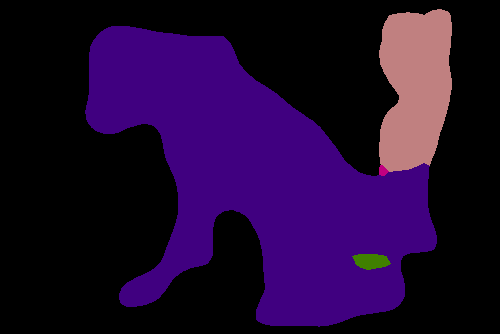} &
			\addFig{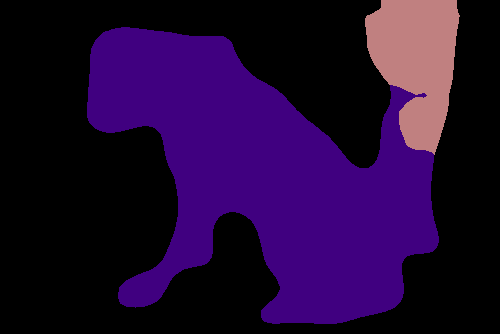} &
			\addFig{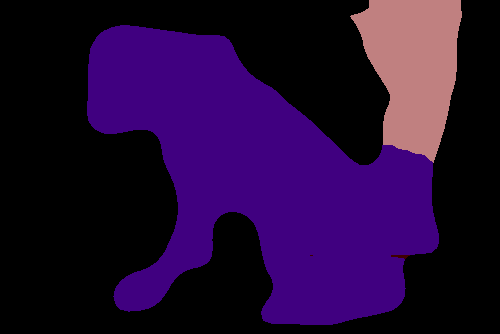} \\
			\addFig{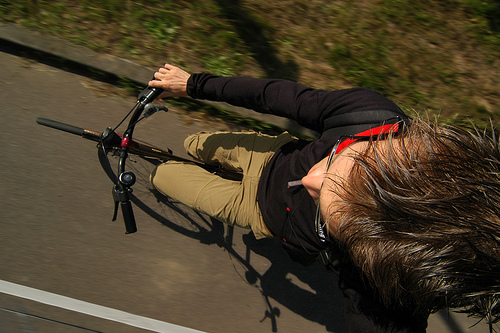} & 
			\addFig{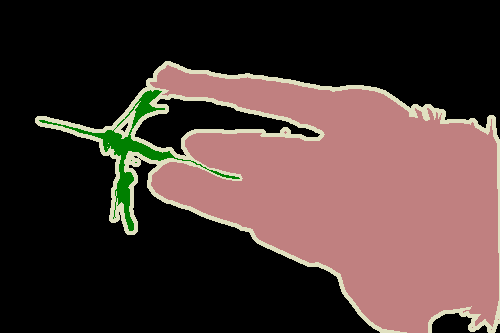} &
			\addFig{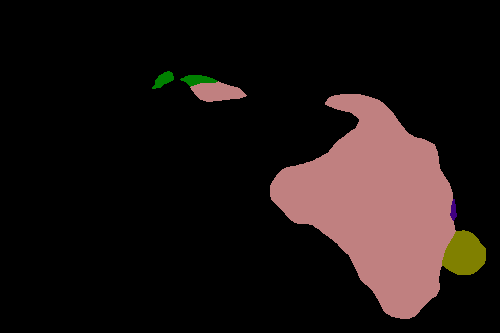} &
			\addFig{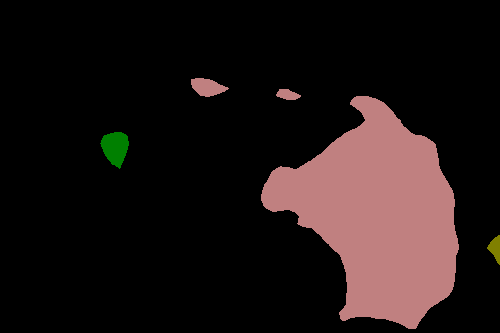} &
			\addFig{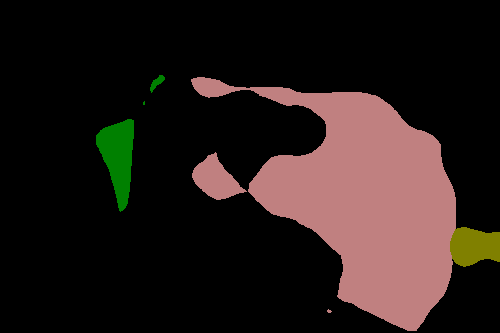} \\
			\addFig{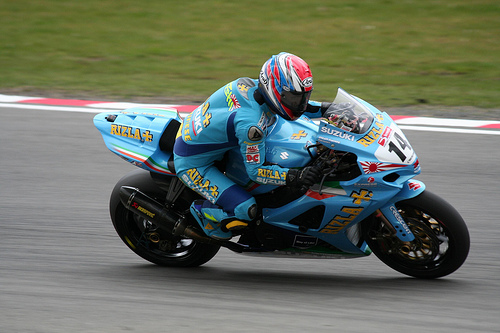} & 
			\addFig{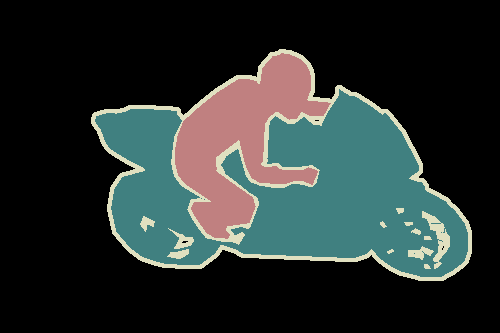} &
			\addFig{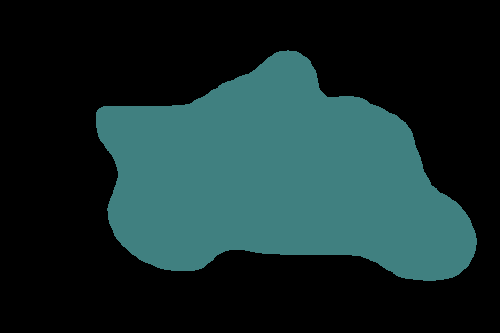} &
			\addFig{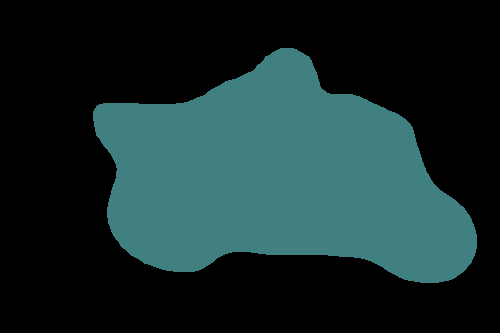} &
			\addFig{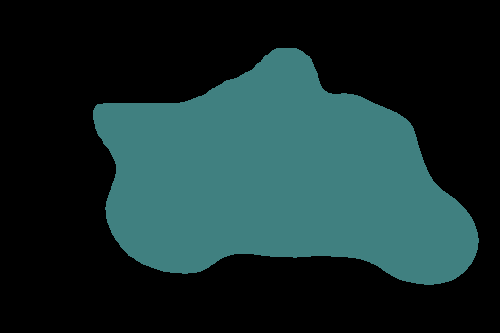} \\
			\addFig{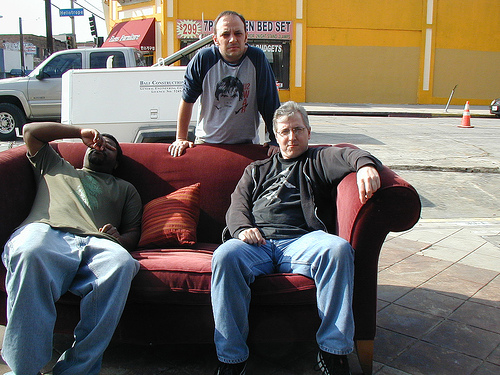} & 
			\addFig{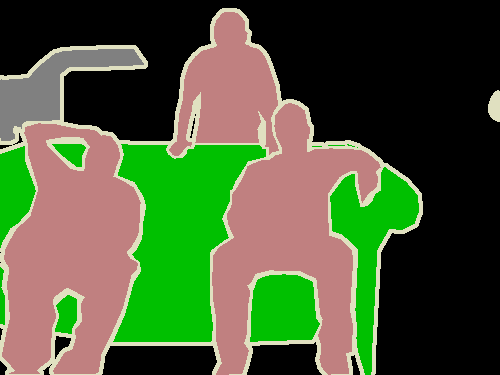} &
			\addFig{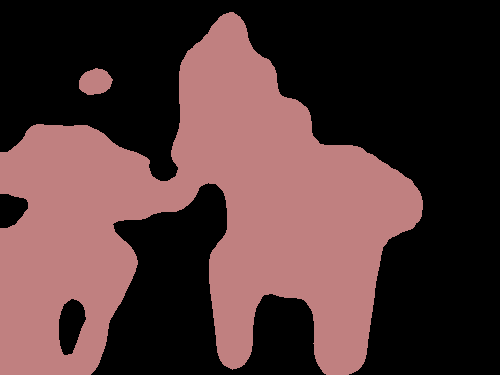} &
			\addFig{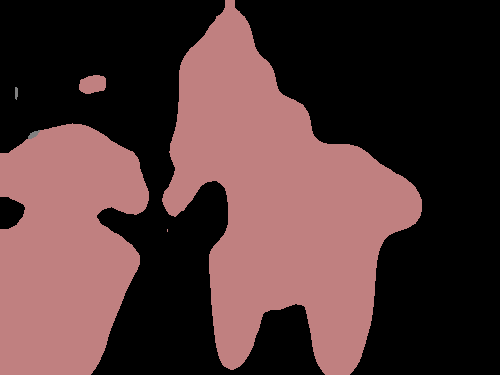} &
			\addFig{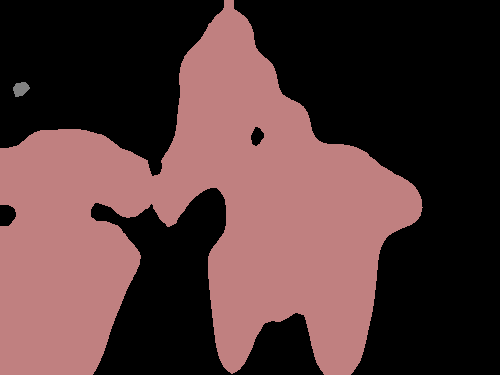} \\
			(1) Source & (2) Ground truth & (3) Initial model & (4) EM 1st iteration &
			(5) EM 2nd iteration
		\end{tabular*}
	}
	\caption{\small \linespread{0.7} Qualitative results for weakly-supervised semantic segmentation using our proposed EM-based method. It is clear that as we move from the initial model (3rd column) to the second iteration of the EM algorithm (5th column), the segmentation quality improves incrementally. The two bottom rows show failure cases. }
	\label{fig:visualComp}
\end{figure*}

\end{document}